\documentclass[10pt]{article} 
\usepackage[preprint]{tmlr}

\usepackage{amsmath,amsfonts,bm}

\def\eqref#1{equation~\ref{#1}}

\def\1{\bm{1}}

\DeclareMathAlphabet{\mathsfit}{\encodingdefault}{\sfdefault}{m}{sl}
\SetMathAlphabet{\mathsfit}{bold}{\encodingdefault}{\sfdefault}{bx}{n}

\usepackage{url}

\usepackage{graphicx}

\usepackage{comment}
\usepackage{amsmath,amssymb} 
\usepackage{color}
\usepackage{booktabs} 
\usepackage{multirow}
\usepackage{array}
\usepackage{tabularx}
\usepackage{mathrsfs}
\usepackage{bm}
\DeclareUnicodeCharacter{2212}{-}
\usepackage[ruled,vlined]{algorithm2e}
\usepackage{caption}
\usepackage{colortbl}
\usepackage{hyperref}
\hypersetup{colorlinks,linkcolor={teal},citecolor={magenta},urlcolor={red}} 
\usepackage[accsupp]{axessibility}  
\usepackage[labelfont=bf]{caption}
\usepackage[nice]{nicefrac}
\usepackage{wrapfig}
\usepackage{paralist}

\definecolor{ForestGreen}{RGB}{34,139,34}
\definecolor{LightCyan}{rgb}{0.88,1,1}

\definecolor{Gray1}{gray}{0.85}
\definecolor{Gray2}{gray}{0.9}
\definecolor{Gray3}{gray}{0.99}
\definecolor{LightCyan}{rgb}{0.88,1,1}
\newcolumntype{a}{>{\columncolor{Gray1}}c}
\newcolumntype{b}{>{\columncolor{Gray2}}c}
\newcolumntype{d}{>{\columncolor{LightCyan}}c}
\newcolumntype{e}{>{\columncolor{Gray3}}c}

\usepackage[symbol]{footmisc}

\title{Uncertainty-aware Test-Time Training (\texttt{UT$^3$}) for Efficient \\ On-the-fly Domain Adaptive Dense Regression}

\author{
    \name Uddeshya Upadhyay \centering 
}

\def\month{MM}  
\def\year{2023} 
\def\openreview{\url{https://openreview.net/forum?id=XXXX}} 

\begin{document}

\maketitle

\begin{abstract}
\vspace{-15pt}
Deep neural networks (DNNs) are increasingly being used in autonomous systems. However, DNNs do not generalize well to domain shift.
Adapting to a continuously evolving environment is a safety-critical challenge inevitably faced by all autonomous systems deployed to the real world. Recent work on test-time training proposes methods that adapt to a new test distribution on the fly by optimizing the DNN model for each test input using self-supervision. However, these techniques result in a sharp increase in inference time as multiple forward and backward passes are required for a single test sample (for test-time training) before finally making the prediction based on the fine-tuned features. This is undesirable for real-world robotics applications where these models may be deployed to resource constraint hardware with strong latency requirements.
In this work, we propose a new framework (called \texttt{UT$^3$}) that leverages test-time training for improved performance in the presence of continuous domain shift while also decreasing the inference time, making it suitable for real-world applications.
Our method proposes an uncertainty-aware self-supervision task for efficient test-time training that leverages the quantified uncertainty to selectively apply the training leading to sharp improvements in the inference time while performing comparably to standard test-time training protocol. Our proposed protocol offers a continuous setting to identify the selected keyframes, allowing the end-user to control how often to apply test-time training.
We demonstrate the efficacy of our method on a dense regression task -- monocular depth estimation. Our experiments show that the standard test-time training technique can be made more efficient via multiple strategies for continuous domain adaptation, and our proposed method, \texttt{UT$^3$}, is the fastest (i.e., more efficient) among all methods in the presence of test data from shifted domain.
\end{abstract}

\vspace{-15pt}
\section{Introduction}
\vspace{-5pt}
Recent advances in deep learning have enabled  a new generation of autonomous agents that can operate in the real world.
These advances were primarily driven by deep learning techniques that have led to breakthroughs for tasks like semantic segmentation~\citep{chen2017deeplab}, monocular depth estimation~\citep{godard2019digging}, and other image-based tasks.
However, while opening new avenues for autonomous agents in the real world, deep learning is often brittle to even subtle domain shifts.
This raises important questions about the reliability and safety of these systems and constitutes a significant obstacle to deploying these systems at scale.
This brittleness of deep predictors can be explained by the phenomenon of covariate shift:
when deployed to the real world, the data distribution gradually shifts away from the distribution the neural network was trained on, leading to undefined behavior.
While the problem could be alleviated through periodic retraining, 
this solution is time-consuming and expensive in both compute and annotation costs.
An alternative is given by domain adaptation~\citep{bridle1990recnorm,ben2010theory}, where the model is trained with labeled data from a source domain and an unlabeled target domain.
However, this solution is still inherently offline and does not resolve the problem of covariate shift within a single mission. Recent works tackle the problem of domain adaptation in an online fashion, however, they largely focus on semantic segmentation and/or classification and assume the existence of a class prototype, which is not possible for dense regression. Also, they often require a replay buffer consisting of samples from the training set that increases the resource requirements for the autonomous system that it is deployed on~\citep{panagiotakopoulos2022online,wang2022continual,iwasawa2021test,liu2023deja}.

\begin{figure}[!t]
    \vspace{-15pt}
    \centering
    \includegraphics[width=0.98\textwidth]{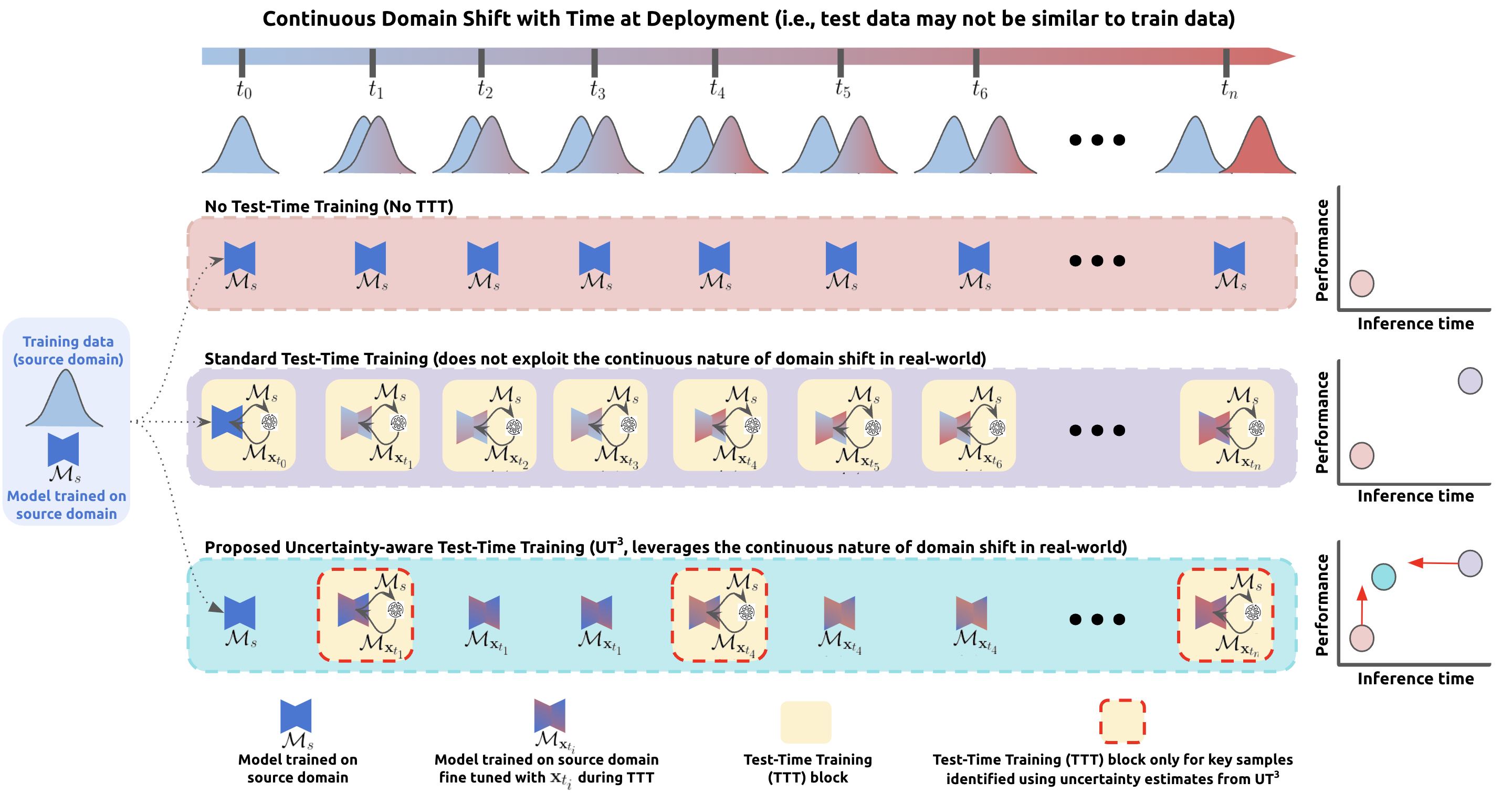}
    \vspace{-5pt}
    \caption{
    In the presence of a continuously drifting domain:
    (Top)~Using a fixed model leads to poor performance. 
    (Middle)~Using \textit{test-time training} leads to improved performance but at higher inference time, limiting real-world use cases.
    (Bottom)~Using \textit{uncertainty-aware test-time training} (\texttt{UT$^3$}) leads to improved performance and reduced inference time, making it applicable to real-world settings.
    }
    \label{fig:arch}
    \vspace{-15pt}
\end{figure}

Recently, Test-Time-Training (TTT)~\citep{sun2020test} was  proposed as a promising alternative to offline domain adaptation.
Test-time training applies a few training steps to the network for each new input using a self-supervised training objective.
As was shown by \citet{sun2020test}, this technique makes neural networks significantly more robust for classification tasks. 
Since its first introduction, multiple self-supervised training objectives have been proposed for test-time training, including entropy minimization~\citep{wang2020tent} and masked autoencoders~\citep{gandelsmantest}.
However, all these techniques have in common that they need to apply multiple training steps (e.g. 30) to each new example at test time, leading to a significant test-time overhead.
In this paper, we address this challenge.
Our key insight is that the test-time overhead can be significantly ameliorated by leveraging advances in uncertainty estimation~\citep{kendall2017uncertainties,upadhyay2022bayescap} for deep neural predictors and by exploiting the continuity of the covariate shift over time.
This is visualized in Fig.~\ref{fig:arch}: 
instead of retraining the network from scratch for every test sample, we propose to intelligently select keyframes, to which we apply test-time-training and preserve the state of the DNNs, till we hit the next keyframe.
We select these keyframes using uncertainty estimates and name our approach \textit{Uncertainty-Aware Test-Time Training}~(\texttt{UT$^3$}).
In addition, we propose a novel uncertainty-aware self-supervised training objective for test-time training of monocular depth estimation and show that test-time training can be effectively applied to a regression problem.
In contrast to classification and segmentation tasks that were studied before, for depth estimation, we can exploit structural knowledge about the 3D world and build on existing self-supervised training objectives for offline training~\citep{godard2019digging}.

In summary, our contributions are as follows:
(i)~We show how test-time training can be made more efficient by intelligently selecting keyframes (using uncertainty estimation) and preserving the context between keyframes. 
(ii)~ We show that test-time training can be effectively applied to dense regression tasks like monocular depth estimation.
(iii)~We propose a novel uncertainty-aware self-supervised training objective for test-time training of monocular depth estimation.
(iv)~We provide experimental insights into the advantages and disadvantages of different variants of test-time training for continuous sequences.

\section{Related Work}
\label{sec:related}

\paragraph{Monocular Depth Estimation.}
Scene depth estimation is an important task in computer vision, which enhances the perception and understanding of real three-dimensional scenes leading to a wide range of applications such as robotic navigation, autonomous driving, and virtual reality~\citep{ming2021deep,el2019survey,ye2017self}. Depth estimation can be broadly classified into (i)~active depth estimation  and (ii)~image-based depth estimation. Active depth estimation leverages lasers, structured light, and other reflections on the object surface to obtain depth point clouds, complete surface modeling, and estimate scene depth maps~\citep{ming2021deep,kumar2018near,park2019high} but these methods involve heavy costs of data preparation, labeling, and computing resources~\citep{ming2021deep}. Image-based depth estimations are more attractive as they cut the need for heavy resources while being applicable to a wide range of applications~\citep{godard2019digging,godard2017unsupervised,fu2018deep,hong2022depth}.
However, most of these work train models once with a fixed dataset before real-world deployment, which is bound to fail in a continuously evolving environment that may drift away from the training dataset. Hence, an efficient online domain adaptive technique is essential to make these methods more robust.

\vspace{-10pt}
\paragraph{Domain Adaptation.}
As discussed above, a significant challenge with machine learning methods that are deployed to the real world is the drift in the input domain at the test time (i.e., the target domain may not be similar to the source domain). Domain adaptation techniques transfer knowledge from a source domain to a target domain. This has been widely studied in various applications, including computer vision~\citep{wang2018deep,csurka2017comprehensive}, natural language processing~\citep{jiang2007instance,ramponi2020neural,blitzer2006domain}, and speech recognition~\citep{deng2017universum,latif2022self}.
One well-studied approach for domain adaptation is the use of domain-invariant features~\citep{fernando2013unsupervised}, which aims to align the source and target domains by learning representations that are invariant to the domain shift. This has been further extended to adversarial learning~\citep{ganin2016domain}, where a discriminator is trained to distinguish between the source and target domains while the feature extractor tries to fool the discriminator.
Another popular approach is to use instance-level adaptation~\citep{zhang2021open,liu2020unsupervised}, where samples from the target domain are re-weighted based on their similarity to source domain samples. This has been shown to be effective in scenarios where the target domain has limited annotated data.
Other works like~\citep{mansour2008domain,wang2020learning,sun2015survey} have also explored combining multiple domains, where the goal is to leverage the information from multiple domains to improve the adaptation performance. This has been shown to be particularly effective in scenarios where the source domains are complementary. However, many previous techniques attempt domain adaptation in an offline fashion, which depends on having certain knowledge about the potentially unseen domain. Works such as~\citep{panagiotakopoulos2022online,wang2022continual,iwasawa2021test,liu2023deja} propose an online scheme to perform domain adaptation in the context of semantic segmentation/classification. However, they assume the existence of a class prototype, which is not possible for dense regression.

\vspace{-10pt}
\paragraph{Test-Time Training (TTT).}
It is a technique where the model is fine-tuned on new data at test time using self-supervision~\citep{sun2020test,liu2021ttt++}. While it attempts to tackle the same problem of domain shift as domain adaptation, the difference is that TTT does not require any information about the potential target SHIFT During training, unlike domain adaptation.
A typical TTT framework consists of a primary-task branch and a self-supervision task branch, both sharing a common encoder. For a sample at test time, first, the encoder and self-supervision head are fine-tuned on the sample. Afterward, the fine-tuned encoder is used with the primary task head to make the final prediction~\citep{lin2022video,gandelsmantest,sun2020test}. This approach has been shown to be effective for computer vision tasks~\citep{zhou2022domain,koh2021wilds,wang2022continual,azimi2022self,wangtest,zhang2022auxadapt}. However, TTT comes at the cost of higher inference time, making them unsuitable for low-latency, real-world applications.

\vspace{-10pt}
\paragraph{Uncertainty Estimation.}
It is a crucial task in machine learning, particularly for deep neural networks applied to safety-critical applications~\citep{gal2016uncertainty,kendall2017uncertainties,laves2020well,laves2020calibration,sudarshan2021towards,upadhyay2021robustness,upadhyay2021uncertainty,upadhyay2021uncertaintyICCV}, as it can allow triggering interventions when the model is not confident in the predictions. Uncertainty estimation can be used for various purposes, such as active learning, model selection, and out-of-distribution detection~\citep{lakshminarayanan2016simple,durasov2021masksembles,malinin2018predictive,shapeev2020active,tian2020role}.
Another popular approach for uncertainty estimation in deep neural networks is Bayesian deep learning~\citep{wilson2020bayesian,wang2020survey,abdullah2022review,daxberger2021laplace,gustafsson2020evaluating}. This approach involves modeling the parameters of the network as random variables and estimating their posterior distribution.
In this paper, we leverage these advances in uncertainty estimation to reduce latency for test-time training by using these estimates to intelligently apply test-time training.

\section{Method}
\label{sec:methods}
Section~\ref{sec:probf} describes the problem formulation. The required preliminaries on test-time training and uncertainty estimation are in Section~\ref{sec:T3} and \ref{sec:unc}. 
In Section~\ref{sec:UT3}, we describe the construction of \texttt{UT$^3$} that proposes an uncertainty-aware test-time training scheme for dense regression tasks (like monocular depth estimation) for efficient test-time training.

\subsection{Problem formulation} 
\label{sec:probf}
Let $\mathcal{D}_s = \{(\mathbf{x}_i, \mathbf{y}_i)\}_{i=1}^{N}$ be the training set with pairs from domain $\mathbf{X}_s$ and $\mathbf{Y}$ (i.e., $\mathbf{x}_i \in \mathbf{X}_s, \mathbf{y}_i \in \mathbf{Y}, \forall i$ and $\mathcal{D}_s$ refers to ``source'' dataset), where $\mathbf{X}_s, \mathbf{Y}$ are subsets of $\mathbb{R}^m$ and $\mathbb{R}^n$, respectively. 
While our formulation is valid 
for arbitrary dimension, we present the formulation for image-based dense regression task: monocular depth estimation~\citep{godard2019digging,godard2017unsupervised,patil2022p3depth}. 
Therefore, ($\mathbf{x}_i, \mathbf{y}_i$) represents a pair of images, where $\mathbf{x}_i$ refers to the input and $\mathbf{y}_i$ denotes the corresponding output. 
For instance, in monocular depth estimation $\mathbf{x}_i$ is an RGB image, and $\mathbf{y}_i$ is the depth map for that image.
Oftentimes, it is of great interest to learn a deep neural network (DNN), $\mathbf{\Psi}(\cdot; \Theta)$, that learns a mapping from $\mathbf{X}_s$ to $\mathbf{Y}$, i.e., $\mathbf{\Psi}(\cdot; \Theta): \mathbf{X}_s \rightarrow \mathbf{Y}$. In this work, we examine the problem of making $\mathbf{\Psi}$ generalizable to an arbitrary shifted domain $\mathbf{X}_t$~\citep{zhao2019geometry,roussel2019monocular}.
For instance, depth estimation models for autonomous driving may be trained in a particular environment but should reliably work in novel environments and surroundings (i.e., different weather conditions, lighting, cities, etc.) encountered during deployment.

\subsection{Preliminaries}
\label{sec:prelim}
\subsubsection{Test-time Training}
\label{sec:T3}
\begin{wrapfigure}{l}{0.35\textwidth}
  \begin{center}
    \includegraphics[width=0.35\textwidth]{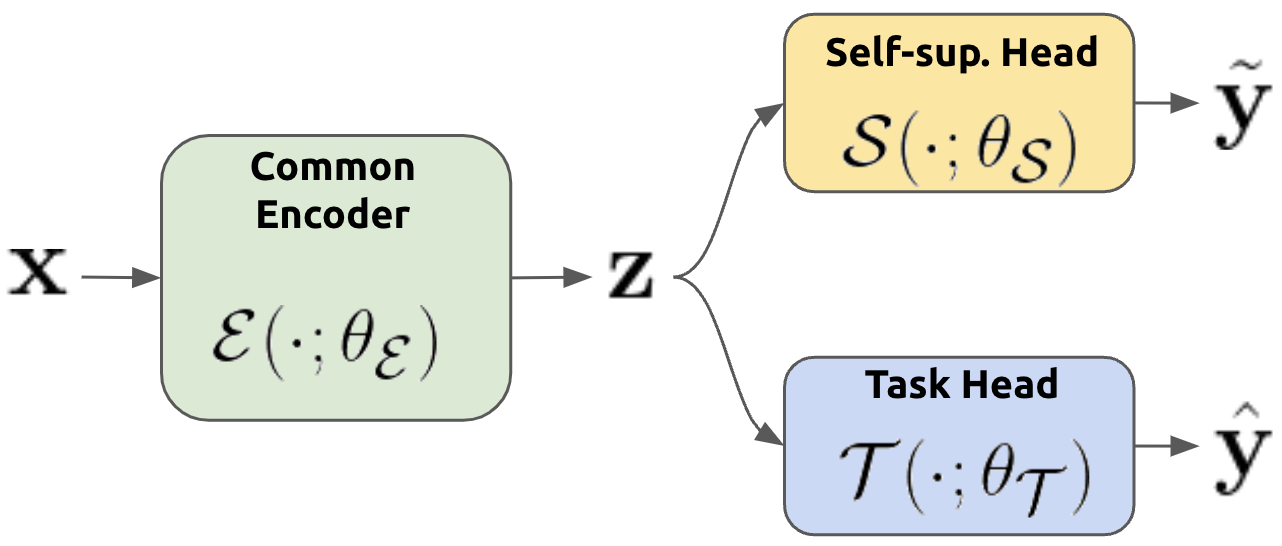}
  \end{center}
  \vspace{-10pt}
  \caption{
  The encoder, task-head, and self-supervision head for the test-time training framework.
  }
  \vspace{-8pt}
  \label{fig:T3}
\end{wrapfigure}
Recent works~\citep{sun2020test,gandelsmantest} have highlighted a general approach for improving the performance of predictive models when training and test data come from different distributions by turning a single unlabeled test sample into a self-supervised learning problem. At each step, the above methods update the model parameters before making a prediction. 
The approach consists of three crucial components: 
(i)~The common encoder $\mathbf{\mathcal{E}}(\cdot; \theta_{\mathcal{E}})$ that maps an input $\mathbf{x}$ to a latent representation $\mathbf{z}$. 
(ii)~The task dependent head $\mathbf{\mathcal{T}}(\cdot; \theta_\mathcal{T})$ that makes the task prediction based on the latent representation $\mathbf{z}$.
(iii)~The self-supervisory head $\mathbf{\mathcal{S}}(\cdot; \theta_\mathcal{S})$ that performs a self-supervision task given a latent representation $\mathbf{z}$.
The three components are arranged in a Y shape with the encoder feeding into both heads as shown in Figure~\ref{fig:T3}.
The parameters for the different modules ($\{\theta_\mathcal{E}^*, \theta_\mathcal{S}^*, \theta_\mathcal{T}^*\}$) can be obtained by solving the following optimization problem which is a combination of self-supervision and task-dependent objectives as described in~\citep{gandelsmantest},
\begin{gather}
    \{\theta_\mathcal{E}^*, \theta_\mathcal{S}^*, \theta_\mathcal{T}^*\} = \underset{\theta_\mathcal{E}, \theta_\mathcal{S}, \theta_\mathcal{T}}{\text{argmin}}
    \left[
    \lambda_1 \underbrace{\mathcal{L}_{ss}(\mathbf{\mathcal{S}}(\mathbf{\mathcal{E}}(\mathbf{x}; \theta_\mathcal{E}); \theta_\mathcal{S}))}_{\text{Self-supervision loss}}
    +
    \lambda_2 \underbrace{\mathcal{L}_{T}(\mathbf{\mathcal{T}}(\mathbf{\mathcal{E}}(\mathbf{x}; \theta_\mathcal{E}); \theta_\mathcal{T}), \mathbf{y})}_{\text{Task-dependent loss}}
    \right]
\end{gather}
Where, $(\lambda_1, \lambda_2)$ are hyperparameters controlling the relative contributions of the two losses. 
While there are other self-supervision techniques for TTT, we use the one
described in~\citep{gandelsmantest} with masked-autoencoders as the self-supervision task leading to a reconstruction based  $\mathcal{L}_{ss}$, i.e.,
$\mathcal{L}_{ss}(\cdot) = ||\mathbf{\mathcal{S}}\left(\mathbf{\mathcal{E}}(\tilde{\mathbf{x}}; \theta_\mathcal{E}); \theta_\mathcal{S}\right) - \mathbf{x}||^2$. Where $\tilde{\mathbf{x}} := \text{\texttt{mask}}(\mathbf{x})$ represents the masked input image. 
$\mathcal{L}_T$ represents the task-dependent loss term (e.g., cross-entropy for classification).
The predictions from the above framework are obtained by a two-step procedure. For a given input $\mathbf{x}$, we perform

\begin{itemize}
    \item[Step 1 (Test-time Training): ] $\theta_{\mathcal{E}, \mathbf{x}}^*, \theta_{\mathcal{S}, \mathbf{x}}^* = \underset{\parbox{2cm}{\centering \small $\theta_\mathcal{E}, \theta_\mathcal{E}$ \\ \tiny for $Q$ steps}}{\text{argmin}}
    \left[
    \mathcal{L}_{ss}(\mathbf{\mathcal{S}}(\mathbf{\mathcal{E}}(\tilde{\mathbf{x}}; \theta_\mathcal{E}); \theta_\mathcal{S}))
    \right]$ with ($\theta_\mathcal{E}, \theta_\mathcal{S}$) init. to ($\theta_\mathcal{E}^*, \theta_\mathcal{S}^*$)
    \item[Step 2 (Prediction): ] $\hat{\mathbf{y}} = \mathbf{\mathcal{T}}(\mathbf{\mathcal{E}}(\mathbf{x}; \theta_{\mathcal{E}, \mathbf{x}}^*); \theta_\mathcal{T}^*)$
\end{itemize}

\subsubsection{Uncertainty Estimation}
\label{sec:unc}
Various works~\citep{gal2016uncertainty,kendall2017uncertainties,lakshminarayanan2016simple,upadhyay2021robustness,laves2020well,rangnekar2023usim} have proposed different methods to model the uncertainty estimates in the predictions made by DNNs for different tasks. Uncertainty can broadly be categorized into two types (i)~aleatoric (uncertainty due to inherent randomness in the data distribution) and (ii)~epistemic (uncertainty due to parameters of the model) uncertainty~\citep{gal2016uncertainty,kendall2017uncertainties}. Interestingly recent works~\citep{kendall2017uncertainties,laves2020well,upadhyay2022bayescap,upadhyay2023hypuc,upadhyay2023probvlm,upadhyay2023likelihood} have shown that for many real-world computer vision applications with access to large datasets, modeling the aleatoric uncertainty allows for capturing erroneous predictions that may happen with out-of-distribution samples. 
In the context of regression, this is done by modeling the residual between the prediction and the ground truth as a parametric distribution (often Gaussian)~\citep{kendall2017uncertainties,laves2020well,sudarshan2021towards}.
Consider a regression DNN model $\mathbf{\Phi}(\cdot; \theta): \mathbb{R}^m \rightarrow \mathbb{R}^n$, with a set of trainable parameters given by $\theta$. To capture the \textit{irreducible} (i.e., aleatoric) uncertainty in the output distribution $\mathcal{P}_{Y|X}$, the model estimates the parameters of the distribution. These are then used to maximize the likelihood function. 
That is,
for an input $\mathbf{x}_i$, the model produces a set of parameters representing the output given by, $\{\hat{\mathbf{y}}_i, \hat{\sigma}_i \} := \mathbf{\Phi}(\mathbf{x}_i; \theta)$, that characterizes the distribution 
$\mathcal{P}_{Y|X}(\mathbf{y}; \{\hat{\mathbf{y}}_i, \hat{\sigma}_i \})$, such that 
$\mathbf{y}_i \sim \mathcal{P}_{Y|X}(\mathbf{y}; \{\hat{\mathbf{y}}_i, \hat{\sigma}_i \})$.
The likelihood $\mathscr{L}(\zeta; \mathcal{D}) := \prod_{i=1}^{N} \mathcal{P}_{Y|X}(\mathbf{y}_i; \{\hat{\mathbf{y}}_i, \hat{\sigma}_i \})$ is then maximized in order to estimate the optimal parameters of the network.
With a \textit{heteroscedastic} Gaussian distribution for $\mathcal{P}_{Y|X}$~\citep{kendall2017uncertainties,wang2019aleatoric}, maximizing the likelihood becomes,
\begin{gather}
\theta^* 
= \underset{\theta}{\text{argmax}} \prod_{i=1}^{N} \frac{1}{\sqrt{2 \pi \hat{\sigma}_i^2}} 
e^{-\frac{|\hat{\mathbf{y}}_i - \mathbf{y}_i|^2}{2\hat{\sigma}_i^2}} 
= 
\underset{\theta}{\text{argmin}} \sum_{i=1}^{N} \frac{|\hat{\mathbf{y}}_i - \mathbf{y}_i|^2}{2\hat{\sigma}_i^2} + \frac{\log(\hat{\sigma}_i^2)}{2} \label{eq:scratch_gauss}\\
\text{Uncertainty}(\hat{\mathbf{y}}_i) = \hat{\sigma}_i^2.
\end{gather}

\subsection{Building \texttt{UT$^3$}: Uncertainty-aware Test-Time Training Module}
\label{sec:UT3}
We first consider a framework for the \textit{primary task}. We focus on image-based dense regression, where it is typical to have modules $\mathcal{E}(\cdot; \theta_E)$ that encodes the input image and $\mathcal{T}(\cdot; \theta_T)$ that uses the encoded features to synthesize the desired output.
To make this framework compatible with test-time training, we notice that we only need to add a self-supervision head $\mathcal{S}(\cdot; \theta_{SS})$ in the existing framework such that it takes the encodings produced by $\mathcal{E}$ and performs the self-supervision task. 
For our experiments, we propose to use a modified uncertainty-aware masked-autoencoding as self-supervision task~\citep{gandelsmantest,upadhyay2022bayescap} where the network $\mathcal{S}(\cdot; \theta_{SS})$ is designed to have a split head to produce a tuple $(\tilde{\mathbf{y}}, \tilde{\mathbf{\sigma}}) := (\left[ \mathcal{S}(\cdot; \theta_{SS}) \right]_{\tilde{\mathbf{y}}}$, $\left[ \mathcal{S}(\cdot; \theta_{SS}) \right]_{\tilde{\mathbf{\sigma}}})$ as described in~\citep{kendall2017uncertainties,laves2020well}. 
Given an input $\mathbf{x}_i$, the encoder $\mathcal{E}$ generates two features: (i)~$\mathcal{E}(\mathbf{x}_i; \theta_E)$, and (ii)~$\mathcal{E}(\tilde{\mathbf{x}}_i; \theta_E)$, with $\tilde{\mathbf{x}}_i = \text{\texttt{mask}}(\mathbf{x}_i)$. The feature $\mathcal{E}(\mathbf{x}_i; \theta_E)$ is passed to the primary task head $\mathcal{T}$ to predict $\hat{\mathbf{y}}_i := \mathcal{T}(\mathcal{E}(\mathbf{x}_i; \theta_E); \theta_T)$ and the feature $\mathcal{E}((\tilde{\mathbf{x}}_i); \theta_E)$ is passed to self-supervision head $\mathcal{S}$ to predict a tuple
$(\tilde{\mathbf{y}}_i, \tilde{\mathbf{\sigma}}_i ) := ( \left[ \mathcal{S}(\mathcal{E}(\tilde{\mathbf{x}}_i; \theta_E); \theta_{SS}) \right]_{\tilde{\mathbf{y}}}, \left[ \mathcal{S}(\mathcal{E}(\tilde{\mathbf{x}}_i; \theta_E); \theta_{SS}) \right]_{\tilde{\mathbf{\sigma}}} )$. At training, all the components are trained jointly by optimizing:
\begin{gather}
    \{\theta_E^*, \theta_{SS}^*, \theta_T^*\} = \underset{\theta_E, \theta_{SS}, \theta_T}{\text{argmin}}
    \frac{1}{N} \sum_{i=1}^{N}
    \left[
    \lambda_1 
    \underbrace{\mathcal{L}_{uSS}(\mathbf{x}_i, \theta_E, \theta_{SS})}_{\text{Unc. self-sup. loss}}
    +
    \lambda_2 \underbrace{\mathcal{L}_{T}(\mathbf{x}_i, \theta_E, \theta_T, \mathbf{y}_i)}_{\text{Task-dependent loss}}
    \right]
    \label{eq:UT3_train}
    \\
    \mathcal{L}_{uSS}(\mathbf{x}_i, \theta_E, \theta_{SS}) = \frac{|\left[ \mathcal{S}(\mathcal{E}(\tilde{\mathbf{x}}_i; \theta_E); \theta_{SS}) \right]_{\tilde{\mathbf{y}}} - \mathbf{x}_i|^2}
    {2 \left[ \mathcal{S}(\mathcal{E}(\tilde{\mathbf{x}}_i; \theta_E); \theta_{SS}) \right]_{\tilde{\mathbf{\sigma}}}^2 }
    +
    \frac{\log \left[ \mathcal{S}(\mathcal{E}(\tilde{\mathbf{x}}_i; \theta_E); \theta_{SS}) \right]_{\tilde{\mathbf{\sigma}}}^2}{2}
\end{gather}
In the above, $\mathcal{L}_{uSS}$ represents the uncertainty-aware self-supervision loss for the masked-autoencoding task. The term $\mathcal{L}_{T}$ is the task-dependent supervised loss term discussed in depth in Section~\ref{sec:UT3mono}. 
The above optimization with an initial dataset $\mathcal{D}_s^{\text{train}} = \{(\mathbf{x}_i, \mathbf{y}_i)\}_{i=1}^{N}$ leads to an optimal set of parameters $\{\theta_E^*, \theta_{SS}^*, \theta_T^*\}$.
At test time, for every input sample $\mathbf{x}$ we first fine-tune the network $\mathcal{E}(\cdot; \theta_E^*)$ and $\mathcal{S}(\cdot; \theta_{SS}^*)$ using the objective function $\mathcal{L}_{uSS}(\mathbf{x}, \theta_E, \theta_{SS})$ for $Q$ steps, leading to new set of parameters $\{ \theta_{E, \mathbf{x}}^*, \theta_{SS, \mathbf{x}}^* \}$. The fine-tuned encoder $\mathcal{E}(\cdot; \theta_{E, \mathbf{x}}^*)$ is used with the primary task-head $\mathcal{T}(\cdot; \theta_T^*)$ to make final prediction 
$\hat{\mathbf{y}} = \mathcal{T}(\mathcal{E}(\mathbf{x}; \theta_{E, \mathbf{x}}^*); \theta_T^*)$.

We note that an inherent limitation of standard TTT approach~\citep{gandelsmantest,sun2020test} is the time increase to make a prediction: For a test sample, there are $Q$ forward-backward passes involved, followed by one final forward pass. One of our key insights is that TTT can be made more efficient for many real-world applications, e.g., monocular depth estimation, by leveraging the fact that test samples are often coming in as a data stream and domain shifts are continuous in nature. That is, the target distribution will not drift  drastically over a short period of time. Therefore, it may not be necessary to perform test-time training on every test sample if we are able to identify the few \textit{key} test inputs on which we perform test-time training, and keep the fine-tuned parameters of the following frames (i.e., preserving the states). We propose an entropy-based criterion to identify such key inputs (also called \textit{keyframes}).

For each sample in the validation dataset from the same domain $\mathcal{D}_s^{\text{val}} = \{(\mathbf{x}_i, \mathbf{y}_i)\}_{i=1}^{M}$, we obtain the $\{ \hat{\mathbf{y}}_i, \tilde{\mathbf{y}}_i, \tilde{\mathbf{\sigma}}_i \}_{i=1}^{M}$ and the entropy of the Gaussian distribution predicted by the self-supervision head per sample (parameterized by $\{ \tilde{\mathbf{y}}_i, \tilde{\mathbf{\sigma}}_i \}$, i.e., $\mathcal{N}(\tilde{\mathbf{y}}_i, \tilde{\mathbf{\sigma}}_i)$). The set of all the entropy values for samples in $\mathcal{D}_s^{\text{val}}$, say 
$\{ S_i = \mathcal{H}(\mathcal{N}(\tilde{\mathbf{y}}_i, \tilde{\mathbf{\sigma}}_i)) \}_{i=1}^{M}$,
are used to design a threshold given by,
\begin{gather}
    \tau_{q, S} = \texttt{quantile}(\{S_i\}_{i=1}^{M}, q)
\end{gather}
Where $q \in [0,1]$ controls how high the threshold is. At test time, we perform test-time training for samples whose entropy is greater than $\tau_{q, S}$. As we will see later (Section~\ref{sec:eff_t3}), this allows us to find the right regime to achieve a target performance with a controlled impact on computational performance.
That is, for a given test sample $\mathbf{x}$ we compute,
\begin{gather}
    (\hat{\mathbf{y}}, \tilde{\mathbf{y}}, \tilde{\mathbf{\sigma}}) =
    \left(
    \mathcal{T}(\mathcal{E}(\mathbf{x}; \theta_{E}^*); \theta_T^*),\text{ }
    \left[\mathcal{S}(\mathcal{E}(\mathbf{x}; \theta_{E}^*); \theta_{SS}^*)\right]_{\tilde{\mathbf{y}}},\text{ }
    \left[\mathcal{S}(\mathcal{E}(\mathbf{x}; \theta_{E}^*); \theta_{SS}^*)\right]_{\tilde{\mathbf{\sigma}}}
    \right)
\end{gather}
If $\mathcal{H}(\mathcal{N}(\tilde{\mathbf{y}}, \tilde{\mathbf{\sigma}})) > \tau_{q, S}$ we perform test-time training. Otherwise, we use the prediction $\hat{\mathbf{y}}$ as is.

\subsection{\texttt{UT$^3$} Applied to Monocular Depth Estimation}
\label{sec:UT3mono}
\begin{wrapfigure}{l}{0.40\textwidth}
  \begin{center}
    \includegraphics[width=0.40\textwidth]{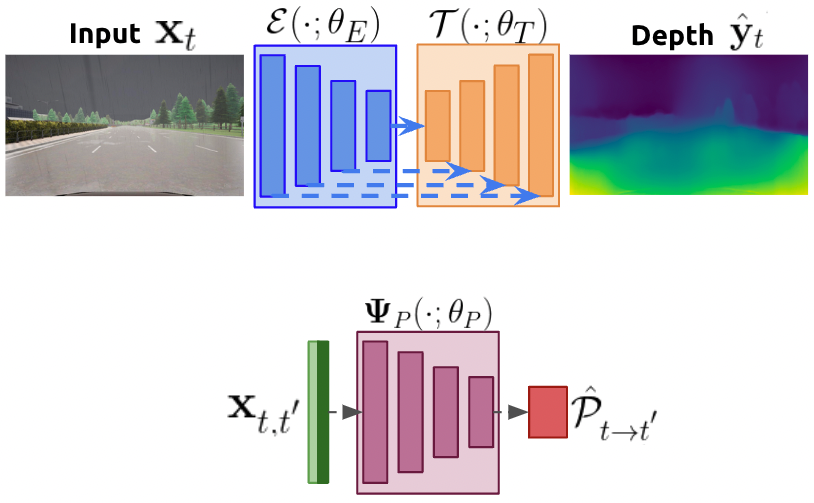}
  \end{center}
  \vspace{-10pt}
  \caption{
    Monodepth2~\citep{godard2019digging} components. It consists of an encoder, a decoder, and a pose-estimator.
  }
  \vspace{-15pt}
  \label{fig:monodepth2}
\end{wrapfigure}
The work in \citep{godard2019digging} proposed a framework called \textit{MonoDepth2} for monocular depth estimation using deep learning. In addition to our proposed method, a key insight in our work is that we can use the self-supervision loss from \cite{godard2019digging} for test-time training without making any changes (detailed in Section~\ref{sec:comp}).
The framework consists of paired encoder-decoder networks $\mathcal{E}$ and $\mathcal{T}$ as shown in Figure~\ref{fig:monodepth2}-(top). In addition to these, there is also a third DNN that estimates the pose estimate between a pair of frames, shown as $\mathbf{\Psi}_P$ in Figure~\ref{fig:monodepth2}-(bottom). The three networks are trained jointly using a combination of photometric projection loss ($\mathcal{L}_p$) and edge-aware smoothing loss ($\mathcal{L}_s$). The loss components are,
\begin{gather}
    \mathcal{L}_p = \underset{t^{'}}{\text{min}} 
    \left[
    \frac{\alpha}{2}(1 - \text{SSIM}(\mathbf{x}_t, \mathbf{x}_{t^{'} \rightarrow t}))
    + (1 - \alpha)|\mathbf{x}_t - \mathbf{x}_{t^{'}
    \rightarrow t}|
    \right] \\
    \mathcal{L}_s = |\partial_x d_t^*|e^{-|\partial_x \mathbf{x}_t|} + |\partial_y d_t^*|e^{-|\partial_y \mathbf{x}_t|} 
\end{gather}
Where $\mathbf{x}_{t^{'} \rightarrow t} := \mathbf{x}_{t^{'}} \left \langle proj(\hat{\mathbf{y}}_t, \hat{\mathcal{P}}_{t \rightarrow t^{'}}, K) \right \rangle$ with $proj(\cdot)$ are the resulting 2D coordinates of the projected depths $\hat{\mathbf{y}}_t$ in $\mathbf{x}_{t^{'}}$, $\langle \cdot \rangle$ is the sampling operator, $K$ is the intrinsics, and $d_t^*$ is the  mean-normalized inverse depth from~\citep{wang2018learning} to discourage shrinking of the estimated depth. The overall objective function for the training is defined by, $\mathcal{L}_{T} = \mu_1 \mathcal{L}_p + \mu_2 \mathcal{L}_s$, where $(\alpha, \mu_1, \mu_2)$ are set to values as described in~\citep{godard2019digging}.
\textit{MonoDepth2} is trained on the KITTI dataset~\citep{Geiger2012CVPR,godard2019digging}. To apply our proposed \texttt{UT$^3$} framework, we take the pre-trained \textit{MonoDepth2} for monocular depth estimation, i.e., we take $\mathcal{E}(\cdot; \theta_{E, K_0}^*), \mathcal{T}(\cdot; \theta_{T, K_0}^*), \mathbf{\Psi}_P(\cdot; \theta_{P, K_0}^*)$ (where $\theta_{E/T/P, K_0}^*$ indicates the trained weights) and introduce the masked-autoencoding self-supervision head $\mathcal{S}(\cdot; \theta_{SS})$, which is designed similar to $\mathcal{T}(\cdot; \theta_{T})$ except the final layers are split into two estimates the mean and variances of \textit{heteroscedastic} Gaussian as described in Section~\ref{sec:UT3}. With the addition of the new network, the entire framework is trained again, where we initialize $\{\theta_E, \theta_T, \theta_P\}$ with the pre-trained weights $\{\theta_{E, K_0}^*, \theta_{T, K_0}^*, \theta_{P, K_0}^*\}$ and $\theta_{SS}$ randomly. We train all the networks jointly on the KITTI dataset by optimizing Equation~\ref{eq:UT3_train} (described in Section~\ref{sec:UT3}) with $\lambda_1 = 1, \lambda_2 = 10^{-3}$ using Adam~\citep{kingma2014adam} optimizer with an initial learning rate set to $10^{-5}$ and decayed using cosine annealing. This yields new values for the set of weights $\{ \theta_{E, K_1}^*, \theta_{T, K_1}^*, \theta_{P, K_1}^*, \theta_{SS, K_1}^* \}$. As described in Section~\ref{sec:UT3}, we perform test-time training with these weights by fine-tuning the network $\mathcal{E}(\cdot; \theta_{E, K_1}^*)$ and $\mathcal{S}(\cdot; \theta_{SS, K_1}^*)$ using the objective function $\mathcal{L}_{uSS}(\mathbf{x}, \theta_E, \theta_{SS})$ for $Q$ steps leading to the new set of parameters $\{ \theta_{E, K_1, \mathbf{x}}^*, \theta_{SS, K_1, \mathbf{x}}^* \}$. The fine-tuned encoder $\mathcal{E}(\cdot; \theta_{E, K_1, \mathbf{x}}^*)$ is used with the primary task-head $\mathcal{T}(\cdot; \theta_{T, K_1}^*)$ to make final prediction 
$\hat{\mathbf{y}} = \mathcal{T}(\mathcal{E}(\mathbf{x}; \theta_{E, K_1, \mathbf{x}}^*); \theta_{T, K_1}^*)$.

\section{Experiments and Results}
\label{sec:exp}
\begin{figure}[!t]
    \centering
    \includegraphics[width=\textwidth]{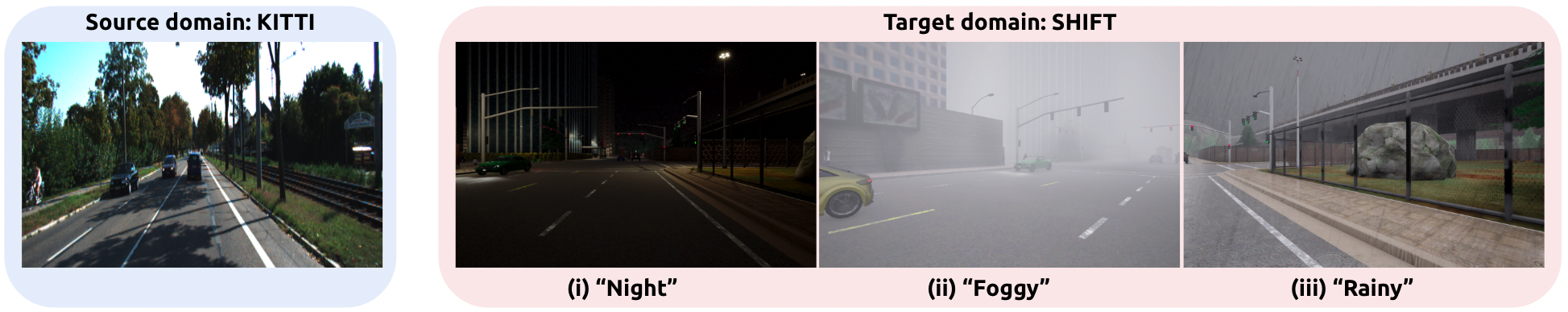}
    \vspace{-15pt}
    \caption{(Left) Sample from the source domain (KITTI). (Right) Samples from target domains (SHIFT)}.
    \label{fig:dset}
    \vspace{-20pt}
\end{figure}
We provide an overview of the experiments performed and the results obtained. In Section~\ref{sec:comp}, we describe the task and various methods used for comparison. Section~\ref{sec:pretrained} analyzes the phenomena of trained model degrading in the presence of data from shifted domains, highlighting the need for adaptation.
Section~\ref{sec:exp_t3} shows that test time training can help significantly improve the performance in the presence of data from continuously shifting domains.
We then present the results demonstrating the improved efficiency of test-time-training methods with the help of uncertainty estimation in Section~\ref{sec:eff_t3}.

\subsection{Tasks, Datasets, and Compared Methods}
\label{sec:comp}
We present the results of all our experiments on the monocular depth estimation task. The base model to perform the task has been trained on the KITTI autonomous driving dataset~\citep{Geiger2012CVPR,godard2019digging}. To evaluate the performance of various methods in the presence of domain shift, we use the SHIFT dataset~\citep{shift2022} for autonomous driving which simulates different weather conditions, lighting conditions, cities, etc, as shown in Figure~\ref{fig:dset}.
We use \textit{MonoDepth2}~\citep{godard2019digging} trained on the KITTI dataset as our base model that performs monocular depth estimation. Following the design principles described in Section~\ref{sec:T3}, we design various test-time training methods based on \textit{MonoDepth2}:
(i)~\texttt{TTT orig-SS} leverages the existing self-supervision component based on photometric reconstruction of \textit{MonoDepth2} to perform test-time training. This involves unfreezing the depth encoder $\mathcal{E}$ and the decoder $\mathcal{T}$ at test time and training them using the loss function described in Section~\ref{sec:UT3mono}.
(ii)~\texttt{TTT MAE} uses an additional self-supervision decoder ($\mathcal{S}$) that is placed on top of the encoder ($\mathcal{E}$) to perform the uncertainty-aware masked autoencoding both at training and test-time training as described in Section~\ref{sec:UT3} \& \ref{sec:UT3mono}. For test-time training, the decoder $\mathcal{T}$ is frozen.
(iii)~\texttt{TTT orig-SS+MAE} refers to the test-time training method that leverages both the original self-supervision technique in the \textit{MonoDepth2} and the additional decoding head to perform masked autoencoding. At test time, all three networks ($\mathcal{E}, \mathcal{T}, \mathcal{S}$) are unfrozen and trained as described in Section~\ref{sec:UT3} \& \ref{sec:UT3mono}.

\subsection{Pre-trained Models on One Domain Perform Poorly on Shifted Domains}
\label{sec:pretrained}
Table~\ref{tab:t1} shows the performance of \textit{MonoDepth2} trained on the KITTI dataset on multiple shifted domains, i.e., ``Night'' (daytime-to-night), ``Foggy'' (clear-to-foggy), ``Rainy'' (clear-to-rainy) as shown in Figure~\ref{fig:dset}.
We see that the pre-trained model performs best on the KITTI dataset (similar domain as that of training domain) with \texttt{Abs Rel} of 0.115. However, the performance of the same model on the other shifted domains degrades substantially with \texttt{Abs Rel} of 0.322 on ``Night'' and 0.301/0.281 on ``Foggy''/``Rainy'' respectively.
Furthermore, we fine-tuned \textit{MonoDepth2} (originally trained on KITTI) on the different SHIFT domains. We observe that the performance of \textit{MonoDepth2} for a particular domain shift improves substantially when we fine-tune the KITTI-based model on that shifted domain; for instance, the performance of a fine-tuned \textit{MonoDepth2} (with ``Night'') on ``Night'' dataset (and not on train dataset) improves with an \texttt{Abs Rel} of 0.263 (compared to 0.322 for non-fine-tuned KITTI-based model). However, the fine-tuned model performs worse on the original KITTI domain (\texttt{Abs Rel} of 0.462).
We observe a similar trend for other SHIFT domains. This indicates that a model trained on a static dataset when deployed to the real world where it may face continuously shifting domains will suffer, and there is a need for methods that can adapt to shifting distributions on-the-fly.
\begin{figure*}[h]
{
\setlength{\tabcolsep}{4pt}
\renewcommand{\arraystretch}{1.2}
\resizebox{\linewidth}{!}{
\begin{tabular}{l|c|a|b|b|b}
  \multirow{1}{*}{\textbf{D}} & \multirow{1}{*}{\textbf{Metrics}} & \textbf{Monodepth-2 (KITTI)} & \textbf{Monodepth-2 (``Night'')} & \textbf{Monodepth-2 (``Foggy'')} & \textbf{Monodepth-2 (``Rainy'')} \\
  \hline
  \hline
  \multirow{6}{*}{\rotatebox[origin=c]{90}{\textbf{KITTI}}}
  & Abs Rel$\downarrow$ & \textbf{0.115} & 0.462 & 0.397 & 0.411  \\
  & RMSE$\downarrow$ & \textbf{4.863} & 6.259 & 5.865 & 6.122  \\
  & RMSE log$\downarrow$ & \textbf{0.193} & 0.497 & 0.523 & 0.561 \\
  & $\delta < 1.25$ $\uparrow$ & \textbf{0.877} & 0.573 & 0.536 & 0.568 \\
  & $\delta < 1.25^2$ $\uparrow$ & \textbf{0.959} & 0.634 & 0.614 & 0.633 \\
  & $\delta < 1.25^3$ $\uparrow$ & \textbf{0.981} & 0.712 & 0.672 & 0.708 \\
  \hline
  \multirow{6}{*}{\rotatebox[origin=c]{90}{\textbf{``Night''}}}
  & Abs Rel$\downarrow$ & 0.322 & \textbf{0.263} & 0.299 & 0.312  \\
  & RMSE$\downarrow$ & 5.742 & \textbf{5.137} & 5.672 & 5.433  \\
  & RMSE log$\downarrow$ & 0.624 & \textbf{0.295} & 0.387 & 0.452 \\
  & $\delta < 1.25$ $\uparrow$ & 0.522 & \textbf{0.711} & 0.646 & 0.632 \\
  & $\delta < 1.25^2$ $\uparrow$ & 0.720 & \textbf{0.772} & 0.733 & 0.754 \\
  & $\delta < 1.25^3$ $\uparrow$ & 0.814 & \textbf{0.836} & 0.831 & 0.822 \\
  \hline
  \multirow{6}{*}{\rotatebox[origin=c]{90}{\textbf{``Foggy''}}}
  & Abs Rel$\downarrow$ & 0.301 & 0.288 & \textbf{0.203} & 0.294  \\
  & RMSE$\downarrow$ & 7.283 & 6.972 & \textbf{4.923} & 6.832  \\
  & RMSE log$\downarrow$ & 0.796 & 0.622 & \textbf{0.223} & 0.654 \\
  & $\delta < 1.25$ $\uparrow$ & 0.598 & 0.664 & \textbf{0.749} & 0.682 \\
  & $\delta < 1.25^2$ $\uparrow$ & 0.749 & 0.763 & \textbf{0.836} & 0.754 \\
  & $\delta < 1.25^3$ $\uparrow$ & 0.806 & 0.812 & \textbf{0.882} & 0.803 \\
  \hline
  \multirow{6}{*}{\rotatebox[origin=c]{90}{\textbf{``Rainy''}}}
  & Abs Rel$\downarrow$ & 0.281 & 0.272 & 0.268 & \textbf{0.197}  \\
  & RMSE$\downarrow$ & 6.681 & 6.255 & 6.517 & \textbf{4.611}  \\
  & RMSE log$\downarrow$ & 0.705 & 0.687 & 0.662 & \textbf{0.239} \\
  & $\delta < 1.25$ $\uparrow$ & 0.628 & 0.693 & o.682 & \textbf{0.788} \\
  & $\delta < 1.25^2$ $\uparrow$ & 0.767 & 0.753 & 0.762 & \textbf{0.844} \\
  & $\delta < 1.25^3$ $\uparrow$ & 0.824 & 0.833 & 0.841 & \textbf{0.878} \\
  \hline
\end{tabular}%
}
\vspace{-5pt}
\captionof{table}{
Comparison showing that \textit{MonoDepth2} performs well (metrics in \textbf{bold}) only on the domain it was trained on, i.e., it does not generalize well to data from a shifted domain, indicating the need for online domain adaptation.
}
\label{tab:t1}
}
\vspace{-10pt}
\end{figure*}

\subsection{Test-Time Training for On-the-fly Domain Adaptation Applied to Dense Prediction}
\label{sec:exp_t3}
Addressing the limitations highlighted in the previous section, we modify the original \textit{MonoDepth2} to incorporate the test-time training, leading to methods \texttt{TTT orig-SS}, \texttt{TTT MAE}, and \texttt{TTT orig-SS+MAE} as described in Section~\ref{sec:UT3mono} \& \ref{sec:comp}.
\begin{figure}[!h]
    \centering
    \includegraphics[width=\textwidth]{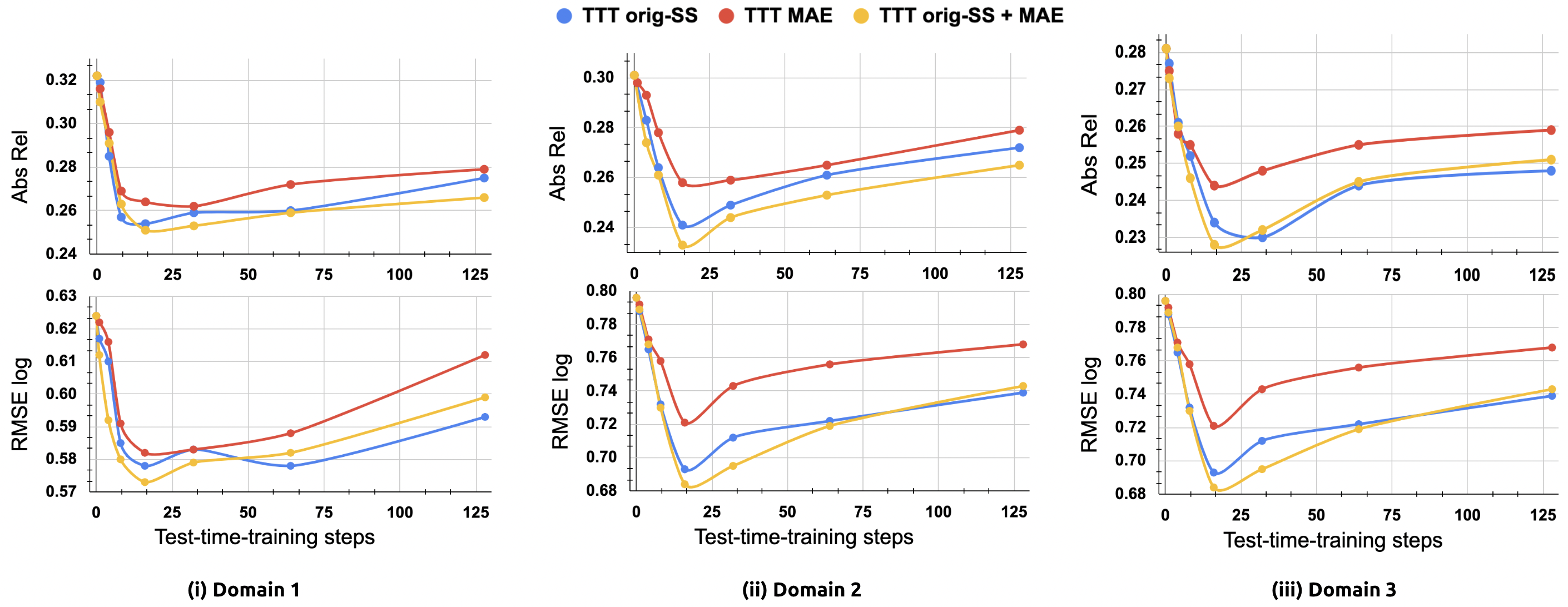}
    \vspace{-20pt}
    \caption{Performance (\texttt{Abs Rel} \& \texttt{RMSE Log}) vs. number of Test-Time 
    Training steps in different shifted domains for methods \texttt{TTT orig-SS}, \texttt{TTT MAE}, and \texttt{TTT orig-SS+MAE}}.
    \label{fig:ttt_metrics}
    \vspace{-15pt}
\end{figure}
Figure~\ref{fig:ttt_metrics} shows the performance of various test-time training techniques on different domains with a varying number of test-time training steps. 
All the models were pre-trained on KITTI and evaluated on the SHIFT dataset~\citep{shift2022} with different domain shifts including Daytime-to-Night (``Night''), Clear-to-Foggy (``Foggy''), Clear-to-Rainy (``Rainy'').
Table~\ref{tab:t2} summarizes the performance of all test-time training methods in terms of different performance metrics in various shifted domains. Figure~\ref{fig:ttt_metrics} shows that \texttt{TTT orig-SS+MAE} performs the best across all domains.
\begin{figure*}[!h]
{
\setlength{\tabcolsep}{4pt}
\renewcommand{\arraystretch}{1}
\resizebox{\linewidth}{!}{
\begin{tabular}{l|l|cccccc}
  \multirow{1}{*}{\textbf{D}} & \multirow{1}{*}{\textbf{Methods}} & \textbf{Abs Rel} $\downarrow$ & \textbf{RMSE} $\downarrow$ & \textbf{RMSE log} $\downarrow$
  & $\delta < 1.25$ $\uparrow$ & $\delta < 1.25^2$ $\uparrow$ & $\delta < 1.25^3$ $\uparrow$\\
  \hline
  \hline
  \multirow{4}{*}{\rotatebox[origin=c]{90}{\tiny\textbf{``Night''}}}
  & \textit{Monodepth2} & 0.322 & 5.742 & 0.624 & 0.522 & 0.720 & 0.814 \\
  &  \qquad + \texttt{TTT orig-SS} & 0.254 & 5.344 & 0.578 & 0.643 & 0.773 & 0.863   \\
  &  \qquad + \texttt{TTT MAE}  & 0.262 & 5.532 & 0.582 & 0.636 & 0.792 & 0.895  \\
  &  \qquad + \texttt{TTT orig-SS+MAE} & \textbf{0.251} & \textbf{5.216} & \textbf{0.573} & \textbf{0.674} & \textbf{0.811} & \textbf{0.907}   \\
  \hline
\multirow{4}{*}{\rotatebox[origin=c]{90}{\tiny\textbf{``Foggy''}}}
  & \textit{Monodepth2} & 0.301 & 7.283 & 0.796 & 0.598 & 0.749 & 0.806  \\
  &  \qquad + \texttt{TTT orig-SS} & 0.241 & 6.874 & 0.693 & 0.635 & 0.773 & 0.833   \\
  &  \qquad + \texttt{TTT MAE} & 0.258 & 7.033 & 0.721 & 0.672 & 0.761 & 0.827   \\
  &  \qquad + \texttt{TTT orig-SS+MAE} & \textbf{0.233} & \textbf{6.722} & \textbf{0.684} & \textbf{0.717} & \textbf{0.794} & \textbf{0.856}   \\
  \hline
  \multirow{4}{*}{\rotatebox[origin=c]{90}{\tiny\textbf{``Rainy''}}}
  & \textit{Monodepth2} & 0.281 & 6.681 & 0.705 & 0.628 & 0.767 & 0.824  \\
  &  \qquad + \texttt{TTT orig-SS} & 0.230 & 6.293 & 0.631 & 0.667 & 0.783 & 0.856  \\
  &  \qquad + \texttt{TTT MAE} & 0.244 & 6.366 & 0.633 & 0.655 & 0.775 & 0.837  \\
  &  \qquad + \texttt{TTT orig-SS+MAE} & \textbf{0.228} & \textbf{6.112} & \textbf{0.618} & \textbf{0.694} & \textbf{0.803} & \textbf{0.886}   \\
  \hline
\end{tabular}%
}
\vspace{-8pt}
\captionof{table}{
Performance comparison for different test-time training techniques on different shifted domains.
}
\label{tab:t2}
}
\vspace{-16pt}
\end{figure*}
For instance, with ``Night'', method \texttt{TTT orig-SS} starts to improve (indicated by lower Abs Rel and RMSE log) as we gradually increase the number of test-time training steps at each frame. It eventually hits the optimal point at close to 16 steps and then performance starts to deteriorate with an increasing number of steps. This phenomenon has been attributed to overfitting in previous works~\citep{gandelsmantest}. While\texttt{ TTT orig-SS} leverages the original self-supervision components in \textit{MonoDepth2} to perform test-time training, a similar trend is observed for \texttt{TTT MAE} that leverages the additional head that performs masked-autoencoding (and does not unfreeze the depth decoder head at test-time). However, the best performance is observed with \texttt{TTT orig-SS+MAE} that combines both the original self-supervision and the additional masked-autoencoding-based self-supervision to perform test-time training.

\begin{wrapfigure}{l}{0.46\textwidth}
  \begin{center}
    \includegraphics[width=0.46\textwidth]{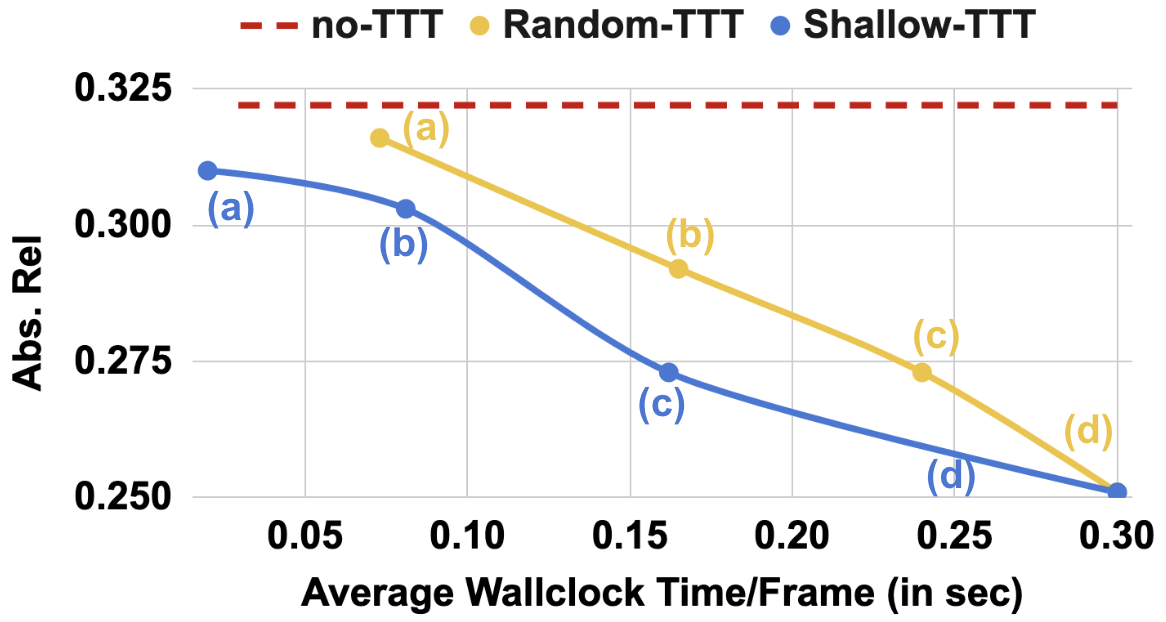}
  \end{center}
  \vspace{-10pt}
  \caption{Making \texttt{TTT orig-SS+MAE} efficient by (i)~applying TTT randomly on some test inputs (\texttt{Random-TTT}), and (ii)~Running TTT for less number of steps (\texttt{shallow-TTT}) on all the test inputs coming from the shifted domain (``Night'').}
  \vspace{-10pt}
  \label{fig:t3_eff}
\end{wrapfigure}
An important point to note is that while test-time training leads to a boost in performance, the multiple forward and backward passes involved during test-time training make it significantly slower at inference time. To make test-time training more efficient, we observe that in the real-world domain, shifts are typically continuous, and test-time training may not be required for all input test samples that are temporally close to each other, as such samples will not have drifted far away from each other in a short time span. As simple baselines, we first investigate two simple strategies for improving inference times which do not leverage temporal continuity. Figure~\ref{fig:t3_eff} shows the performance of the \texttt{TTT orig-SS+MAE} method when it is randomly performing test-time training at different input test instances (represented as \texttt{Random-TTT}). Point (a) for \texttt{Random-TTT} indicates the application of test-time training at any given test input with only 25\% chance of preventing expensive test-time training for many input samples leading to a small average wall-clock time spent per sample, however, the performance gain is also insignificant, compared to no application of TTT as shown with the dashed line. Points (b), (c), and (d) on the \texttt{Random-TTT} curve show the results for applying test-time training to each test input randomly with a probability of 50\%, 75\%, and 100\%. As test-time training is applied to more samples, the performance improves but the average wall-clock time spent per sample also increases.
To strike an optimal balance between wall-clock time at inference and performance, method \texttt{Shallow-TTT} in Figure~\ref{fig:t3_eff} indicates the application of test-time training for every test input but with less number of steps. Points (a), (b), (c), and (d) indicate the application of test-time training with 2, 4, 8, and 16 steps. While shallow test-time training is faster, it does not lead to large gains in performance.

\subsection{Systematically Improving the Efficiency of Test-Time Training}
\label{sec:eff_t3}
As discussed in previous section, while test-time training improves the performance of monocular depth estimation in the presence of domain shifts, it comes at the cost of increased inference time, making it inefficient and undesirable for real-world applications. Figure~\ref{fig:t3_eff} suggests that it may be possible to achieve an optimal balance between performance and inference time by selectively applying test-time training and exploiting the continuous nature of domain shift by preserving the state of models between keyframes. While the experiment with \texttt{Random-TTT} in Figure~\ref{fig:t3_eff} randomly applies test-time training to certain samples, we refine it further by applying test-time training to only certain ``key'' test samples/frames and preserving the state of models between consecutive keyframes. We study the results with a naive strategy and an intelligent strategy to  
to identify ``keyframes'' to apply TTT. Now, we use \texttt{TTT orig-SS+MAE} as  our TTT method.  
\begin{figure}[!h]
    \centering
    \includegraphics[width=\textwidth]{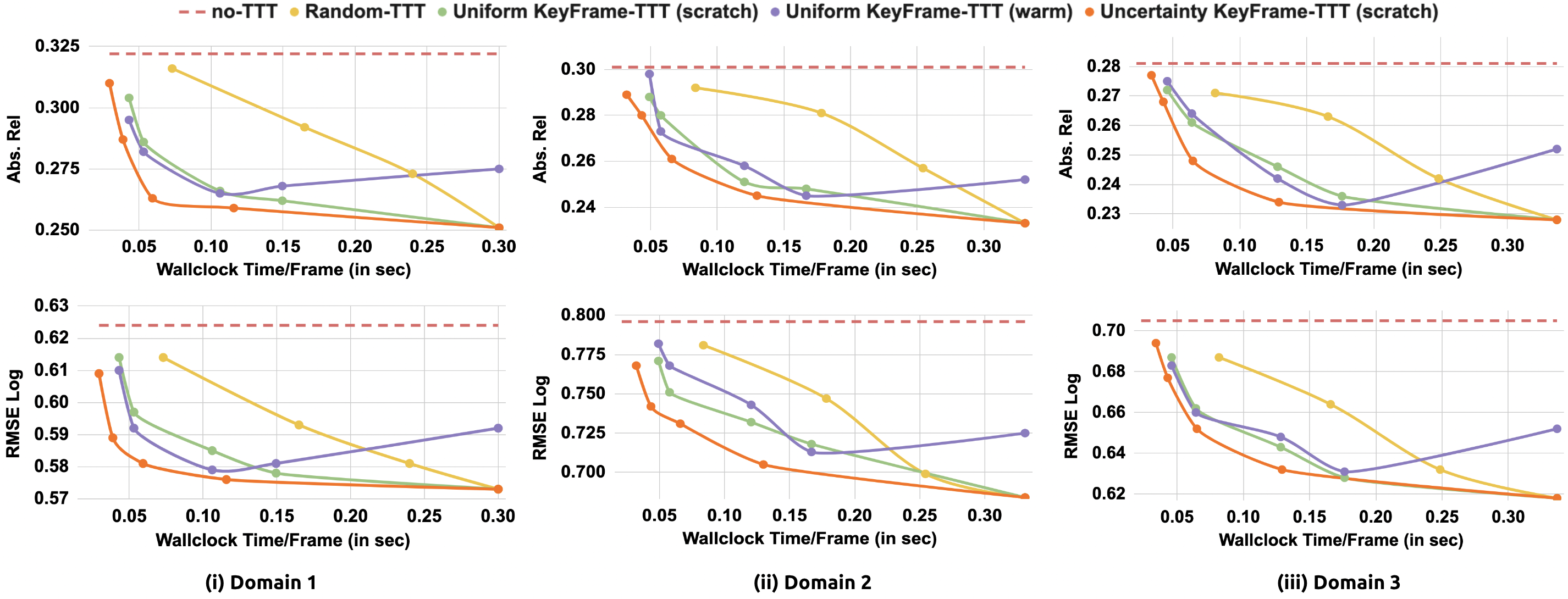}
    \vspace{-15pt}
    \caption{Test-Time Training can be made more efficient (i.e., achieving better performance faster) by selectively applying Test-Time Training only to keyframes. These figures visualize the performance for various strategy to identify keyframes as described in Section~\ref{sec:eff_t3}.}
    \vspace{-18pt}
    \label{fig:wclock}
\end{figure}

\paragraph{\textit{Naive Strategy 1} -- Uniformly Distributed keyframes (scratch) with preserved states.} 
In this strategy, we assume that every $n^{th}$ frame (e.g., every $10^{th}$ frame) in a stream of inputs is a keyframe, and the intermediate frames between two consecutive keyframes do not change drastically. The standard TTT techniques, where test-time training happens at every frame, is a special case of this naive strategy with $n=1$.
Figure~\ref{fig:wclock} shows the results of the experiments for TTT with this strategy for different values of $n \in \{20, 10, 5, 2, 1\}$, referred to as \texttt{Uniform KeyFrame-TTT (scratch)}. Here \texttt{(scratch)} refers to performing TTT at every keyframe starting from the checkpoint of the original source domain. Also, we preserve the state of the models between two consecutive keyframes.
We notice that as the value of $n$ decreases (i.e., TTT is applied to more frames) the performance improves, but the average wallclock time per frame also increases. It can also be observed that this naive strategy performs better than randomly selecting the frames for TTT (i.e., \texttt{Random-TTT}). While effective, it assumes that novel keyframes appear uniformly in the stream, which is not true. As a result, when a smaller number of frames is selected for TTT (i.e., $n=20$), the gain in performance is not significant (compared to not applying TTT), indicating that a better proxy to identify keyframes may lead to improvements.

\paragraph{\textit{Naive Strategy 2} -- Uniformly Distributed keyframes (warm) with preserved states.}
Similar to \textit{Naive Strategy 1}, this strategy assumes that keyframes are uniformly distributed. However, TTT at any given frame uses the checkpoints from the last TTT frame for initialization (unlike \textit{Naive Strategy 1} that uses the checkpoint trained on the source domain).
Figure~\ref{fig:wclock} shows the results of the experiments for TTT with this strategy for different values of $n \in \{20, 10, 5, 2, 1\}$, referred to as \texttt{Uniform KeyFrame-TTT (warm)}.
We notice that in the beginning, as the value of $n$ decreases (i.e., TTT is applied to more frames), performance improves, but after a point when TTT is applied frequently, performance starts to degrade, indicating that using the previous TTT step checkpoints eventually leads to overfitting.

\begin{figure}[!t]
    \centering
    \includegraphics[width=0.99\textwidth]{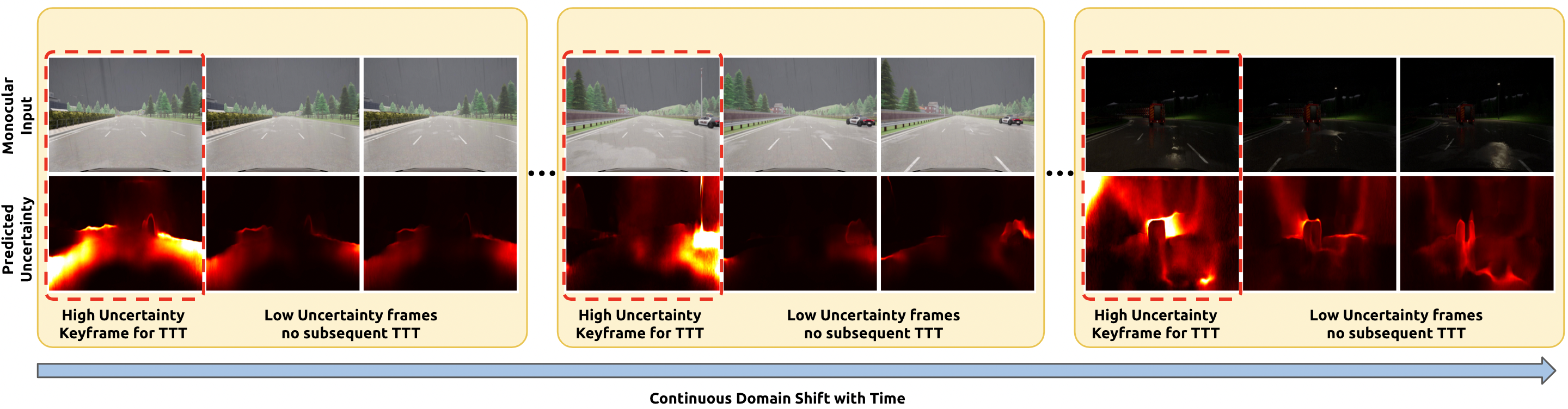}
    \vspace{-5pt}
    \caption{
    Uncertainty-based key-frame identification for selective TTT with state preservation between keyframes (i.e., applying TTT to samples that have higher uncertainty/entropy).
    From left-to-right stream of inputs (top) with the corresponding uncertainty estimates (bottom). The novel frame has higher entropy and is flagged as a keyframe (red-dashed box). The yellow box shows frames for which model states are preserved.
    }
    \label{fig:dshift}
    \vspace{-15pt}
\end{figure}
\paragraph{\textit{Intelligent Strategy} -- Uncertainty keyframes (scratch) with preserved states.}
The above strategies highlight the importance of choosing the right keyframes to apply TTT. 
The proposed \texttt{UT$^3$} framework provides an informative proxy to identify keyframes by quantifying the uncertainty in the reconstructions as described in Section~\ref{sec:UT3} \& \ref{sec:UT3mono}. Given the associated uncertainty estimates for every input frame, we quantify the entropy of the input frame and compare it with the distribution of entropy values for the samples in the source domain. A threshold for the entropy value is defined using the source SHIFT Distribution. At the test time, samples with higher entropy are flagged as keyframes for TTT.
\begin{wrapfigure}{l}{0.35\textwidth}
  \begin{center}
    \includegraphics[width=0.35\textwidth]{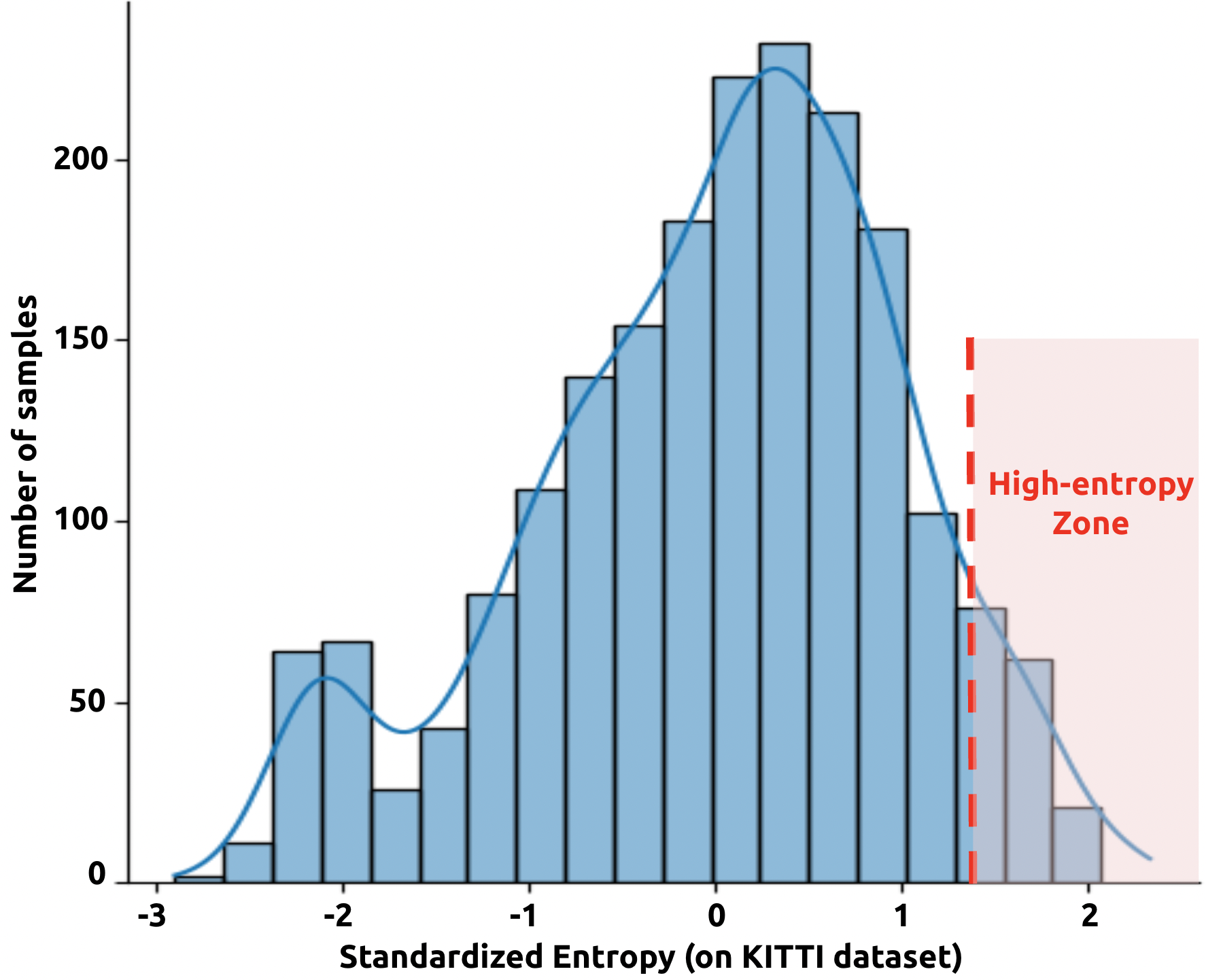}
  \end{center}
  \vspace{-10pt}
  \caption{
    Entropy distribution derived from \texttt{TTT orig-SS+MAE}. 
  }
  \vspace{-10pt}
  \label{fig:ent_dist}
\end{wrapfigure}
Figure~\ref{fig:ent_dist} shows the distribution of standardized entropy values (i.e., entropy $\kappa$ is transformed to $\tfrac{\kappa - \bar{\kappa}_s}{\mathcal{V}_{\kappa, s}}$, where $\bar{\kappa}_s$ is the mean and $\mathcal{V}_{\kappa, s}$ the variance  of all entropy values in source domain) for the samples in source domain that we obtain from the trained \texttt{TTT orig-SS+MAE} method. It also shows the threshold that is set at the $q \%$-tile beyond which is the high-entropy zone (shown in red) where the frames will be flagged as keyframes and TTT will be performed. 
Figure~\ref{fig:dshift} shows a qualitative example of the processing of input frames (belonging to ``Night'') at the test time. It indicates that novel frames (Figure~\ref{fig:dshift}-top row), when first encountered, have high uncertainty values (shown in bright colors, which translates to high entropy) produced by the self-supervision head that performs uncertainty-aware masked-autoencoding (Figure~\ref{fig:dshift}-bottom row). These novel frames are selected as keyframes. After completing TTT on a keyframe, the subsequent frames have lower uncertainty estimates (hence lower entropy) and do not get flagged as key-frame for TTT until the next novel input frame appears.
Figure~\ref{fig:wclock} shows results for this strategy (\texttt{Uncertainty keyframes (scratch)}) with different entropy thresholds (a higher threshold leads to less number of keyframes) in comparison to other strategies. We observe that uncertainty-based key-frame identification can achieve the best performance $\approx 70\%$ faster across all the domains.

\section{Discussion and Conclusion}
In this work, we presented a novel framework named \texttt{UT$^3$} that performs efficient test-time training. The proposed method leverages an uncertainty-aware self-supervision task to quantify the uncertainty (and the entropy) at test time, allowing for the selective application of test-time training. The selective application of test-time training leads to substantial improvements in inference time, making it suitable for applications where limited computing resources and high latency are often tight constraints.
We demonstrated the effectiveness of our method on monocular depth estimation. Our experiments study various strategies using test-time training and show that the proposed method improves the latency-performance trade-off of test-time training in the presence of test data from shifted domains. Our method allows autonomous systems to adapt better to changing environments while reducing the computational burden on the system.
The proposed \texttt{UT$^3$} framework offers a promising solution to the challenge of adapting to continuously evolving environments in autonomous systems by offering a lever to find the right test-time training setup that allows a trade-off between performance and computational efficiency, making it suitable for real-world applications. We believe that this work will pave the way for further advancements in the field of test-time training and provide a more efficient approach for autonomous systems to handle the dynamic nature of the real world.

\bibliography{main}
\bibliographystyle{tmlr}

\end{document}